\documentclass{article} 
\usepackage{iclr2027_conference,times}

\usepackage{amsmath,amsfonts,bm}

\def\eqref#1{equation~\ref{#1}}

\def\1{\bm{1}}

\DeclareMathAlphabet{\mathsfit}{\encodingdefault}{\sfdefault}{m}{sl}
\SetMathAlphabet{\mathsfit}{bold}{\encodingdefault}{\sfdefault}{bx}{n}

\usepackage{hyperref}
\usepackage{url}

\usepackage{amsmath}
\usepackage{multirow}
\usepackage{graphicx}
\usepackage{booktabs}
\usepackage{wrapfig}
\usepackage{longtable}

\title{Just-In-Time Agent Memory with Runtime Agentic Research}

\author{
Bingyu Yan\textsuperscript{1},
Chaofan Li\textsuperscript{1},
Hongjin Qian\textsuperscript{1,2},
Shuqi Lu\textsuperscript{1},
Chaozhuo Li\textsuperscript{1},
Zheng Liu\textsuperscript{1,3}\thanks{Project lead} \\
\textsuperscript{1} Beijing Academy of Artificial Intelligence \\
\textsuperscript{2} Peking University \\
\textsuperscript{3} Hong Kong Polytechnic University \\
\texttt{\{zhengliu1026\}@gmail.com}
}

\iclrfinalcopy 
\begin{document}

\maketitle

\begin{abstract}
Memory is critical for AI agents. Many existing agent-memory systems follow an Ahead-of-Time (AOT) design, constructing memory before a specific request arrives. While this reduces online serving cost, such request-agnostic memory construction can discard fine-grained information that later becomes important. To address this limitation, we propose Just-In-Time Agent Memory (JAM), a trainable framework for query-conditioned context construction at runtime. A Memorizer preserves complete raw histories in a hierarchical page-store with compact navigational summaries, while a Researcher iteratively retrieves, inspects, and integrates evidence for each request. To train these memory-use behaviors, we introduce Memory-Gym, an evidence-grounded data synthesis pipeline covering nine task types across six domains, and optimize the Researcher through verified-trajectory supervised fine-tuning followed by Hint-guided Group Relative Policy Optimization. We demonstrate the effectiveness of JAM across a variety of benchmarks on agent memory and long-context processing, where it achieves stronger task performance than AOT-style memory systems while remaining substantially more efficient than prior trained agentic memory approaches. To support reproducibility and future research, we release our anonymized source code at \url{https://github.com/VectorSpaceLab/general-agentic-memory}.

\end{abstract}

\section{Introduction}

Recent progress in large language models (LLMs) has accelerated the development of AI agents capable of carrying out increasingly complex workflows, such as deep research, software engineering, and other multi-stage applications~\cite{huang2025deep,he2025llm}. 
Such workflows continuously produce long and evolving histories of observations, reasoning traces, and interactions, making agent memory essential for constructing the context needed by downstream tasks~\cite{hu2025memory}.

Existing agent memory systems often follow an Ahead-of-Time (AOT) design philosophy. 
They compress or organize raw histories before any concrete request arrives, producing pre-constructed memories such as summaries or memory notes~\cite{xu2025mem,kang2025memory,fang2025lightmem}. 
Although this request-agnostic preprocessing reduces online serving cost, it can discard fine-grained details and cross-session dependencies that may later become crucial. 
Such information loss is particularly problematic when task-relevant evidence is dispersed across large, evolving, and temporally interdependent histories, rather than contained in a single retrievable passage~\cite{locomo,wu2024longmemeval}. 
Consequently, fixed retrieval or summarization pipelines struggle to assemble the query-specific context needed for downstream reasoning.

These limitations suggest that agent memory should not be viewed merely as storing past information, but as constructing query-relevant context at runtime. 
We therefore propose \textbf{Just-In-Time Agent Memory (JAM)}, a dual-component framework that follows a Just-In-Time (JIT) design philosophy. 
Instead of relying on fully pre-compiled memory, JAM separates offline memory organization from runtime memory construction. 
During the offline stage, a \textit{Memorizer} preserves complete historical records in a page-store while maintaining lightweight memory as navigational guidance. 
At runtime, a \textit{Researcher} conducts runtime agentic research guided by the lightweight memory, retrieving, inspecting, and integrating relevant information to assemble an optimized context for each request. 
By shifting memory from request-agnostic compression to query-conditioned context construction, JAM reduces information loss and improves adaptability to dynamic information needs.

JAM exposes runtime memory construction as an explicit agentic process, making the Researcher's thinking, tool-use, and sufficiency-assessment behaviors traceable and trainable.
Although recent training-based memory agents have explored supervision or reinforcement learning for memory use~\cite{yu2025memagent,yan2025memory,wang2025mem,shi2025look,yue2026mem}, their training data are often derived from conventional multi-hop QA datasets or benchmark-specific splits, offering limited coverage of realistic long-history memory scenarios. 
To address this data scarcity, we introduce \textbf{Memory-Gym}, a scalable data synthesis pipeline that provides evidence-grounded training scenarios for memory agents. 
Memory-Gym covers single-hop localization, multi-hop evidence connection, and multi-session summarization across heterogeneous domains, producing queries, answers, and supporting evidence over accumulated histories.

Building on Memory-Gym, we optimize JAM through a cascaded SFT-RL workflow. 
Supervised fine-tuning (SFT) on verified teacher-generated trajectories first initializes the Researcher's thinking, tool-use, and sufficiency-assessment behaviors. 
Hint-guided Group Relative Policy Optimization (GRPO) then further improves multi-step evidence exploration under sparse feedback, enabling JAM to construct more accurate query-relevant contexts at runtime.

Our contributions are summarized as follows:
(1) We introduce a Just-In-Time paradigm, which shifts agent memory from static Ahead-of-Time compression to runtime query-conditioned context construction.
(2) We propose JAM, a dual-component framework that instantiates this paradigm with a Memorizer for lightweight offline organization and a Researcher for runtime agentic research over complete historical records.
(3) We introduce Memory-Gym, a scalable data synthesis pipeline that provides various training scenarios for memory agents, and develop a cascaded optimization workflow that combines verified-trajectory SFT with Hint-guided GRPO to improve JAM's runtime agentic research.
(4) Extensive experiments demonstrate that JAM consistently improves task performance over AOT-style memory systems while providing a favorable quality--cost trade-off relative to trained agentic memory approaches.

\begin{figure}[t]
    \centering
    \includegraphics[width=\textwidth]{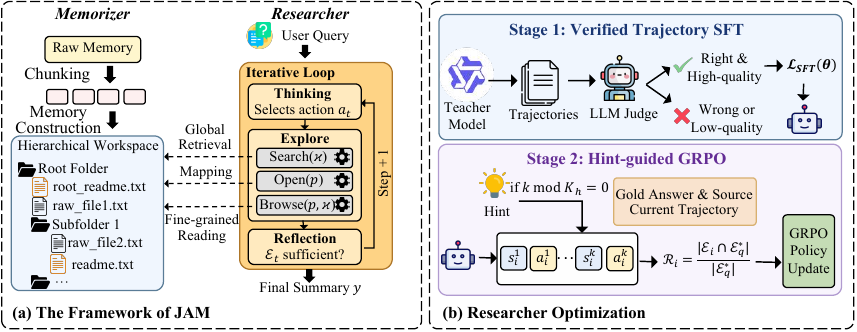} 
    \caption{Overview of JAM. (a) The dual-agent JAM framework, where the Memorizer builds a hierarchical workspace and the Researcher performs iterative memory exploration. (b) The two-stage Researcher optimization pipeline, consisting of verified-trajectory SFT and Hint-guided GRPO.} 
    \label{fig:main} 
\end{figure}

\section{Just-In-Time Agent Memory}

JAM instantiates JIT memory with two components: a Memorizer that preserves
and organizes accumulated histories offline, and a Researcher that
constructs query-conditioned context at runtime.

\subsection{Problem Formulation}

AI agents often accumulate long histories while performing complex tasks such as software engineering and deep research. 
We represent such a history as a sequence of temporally ordered sessions 
$H=\{s_1,s_2,\ldots,s_T\}$, where each session may contain observations, reasoning traces, or interactions. 
As $H$ grows, directly feeding the full history to the agent becomes inefficient and error-prone, motivating a memory system that produces a compact yet sufficient context for each online request.

Given an online request $q$ and an accumulated history $H$, the memory system produces a context for the client agent:
$c^* \leftarrow \mathrm{Memory}(q,H).$
Ideally, this context should be both useful and compact: it should maximize the client agent's downstream performance while minimizing context size or serving cost. 
We formalize this cost-effectiveness objective as:
\begin{equation}
\mathcal{C}^*(q,H)
=
\operatorname*{arg\,max}_{c \in \mathcal{C}(H)}
\mathrm{Perf}_{\mathcal{A}}(q,c),
\qquad
c^*
=
\operatorname*{arg\,min}_{c \in \mathcal{C}^*(q,H)}
|c|.
\end{equation}
Here, $\mathcal{C}(H)$ denotes the space of contexts constructible from history $H$, $\mathcal{C}^*(q,H)$ denotes the set of performance-optimal contexts for request $q$, $\mathrm{Perf}_{\mathcal{A}}(q,c)$ measures the downstream performance of the client agent $\mathcal{A}$ given context $c$, and $|c|$ denotes context size or serving cost. 
Under this formulation, AOT-style memory approximates $c^*$ through request-agnostic pre-compression, whereas JIT memory constructs $c^*$ at runtime conditioned on the specific request $q$.



\subsection{Memorizer: Hierarchical Workspace Construction}

The Memorizer operates during the offline stage, converting accumulated agent histories into a hierarchical workspace that serves as JAM's persistent page-store. 
Given a history $H=\{s_1,\ldots,s_T\}$, it preserves raw sessions while constructing compact navigational summaries through two lightweight operations: \textit{memorizing} and \textit{workspace writing}. 
The same procedure supports both batch construction from existing histories and incremental updates when new sessions are appended.

First, \textit{memorizing} produces a concise memo $\mu_i$ for each session $s_i$, summarizing its key information and serving as a lightweight descriptor for later organization and navigation:
\begin{equation}
    \mathrm{Memorizer.memorize}(s_i, \mathcal{W}_{<i}) \rightarrow \mu_i ,
\end{equation}
where $\mathcal{W}_{<i}$ denotes the workspace context before incorporating $s_i$.

Second, \textit{workspace writing} stores the raw content of $s_i$ as a source file and uses its memo $\mu_i$ to organize the file within the hierarchical workspace:
\begin{equation}
    \mathrm{Memorizer.write}(s_i, \mu_i, \mathcal{W}_{<i}) \rightarrow \mathcal{W}_{\leq i}.
\end{equation}
During this process, semantically related sessions are grouped into coherent directories according to their memos. 
Each directory maintains a README constructed from the memos of its child files and subfolders, providing a compact navigational summary of the corresponding workspace region.

This design keeps raw files as complete, path-traceable evidence, while memos and READMEs guide the Researcher toward relevant sessions for fine-grained inspection.

\subsection{Researcher: Runtime Agentic Research}

During online serving, given a request $q$ and the hierarchical workspace $\mathcal{W}$, the Researcher iteratively explores the workspace, collects useful information, and assesses sufficiency to construct an optimized context for the request.

The Researcher is equipped with three workspace-access tools: 
\texttt{open}$(p)$ reads the README and directory listing under path $p$; 
\texttt{search}$(\kappa)$ performs hybrid retrieval on candidate paths using BM25 and dense embeddings; 
and \texttt{browse}$(p,\kappa)$ reads a specific path and extracts query-relevant information. 
These tools define the action space for runtime exploration.

We formulate the Researcher as an iterative thinking--exploration--reflection process. 
At step $i$, the Researcher observes the request $q$ and the previous interaction trajectory $h_{i-1}$. 
It first performs \textit{thinking} to identify the current information needs and generate a set of memory-access actions:
\begin{equation}
    A_i=\{a_i^1,\ldots,a_i^{K_i}\}
    \sim \pi_{\theta}(\cdot \mid q,h_{i-1}),
    \qquad 
    a_i^k=(t_i^k,\rho_i^k),\quad
    t_i^k \in \mathcal{T},
\end{equation}
where $\mathcal{T}=\{\texttt{open},\texttt{search},\texttt{browse}\}$, $K_i$ is the number of tool calls generated at step $i$, and $\rho_i^k$ denotes the parameters of the selected tool. 

The \textit{exploration} step executes all tool calls in $A_i$ over the workspace $\mathcal{W}$ and returns a set of observations:
$O_i=\{o_i^1,\ldots,o_i^{K_i}\}, \quad o_i^k=\mathrm{Exec}(a_i^k,\mathcal{W}).$
The interaction trajectory is then updated by appending the tool calls and their observations: $ h_i = h_{i-1} \oplus (A_i,O_i).$



After exploration, the Researcher implicitly assesses information sufficiency over the accumulated trajectory:
$y_i = \operatorname{Reflect}_{\theta}(q, h_i)$. It stops when $y_i$ is true or the configured action budget is reached, and continues otherwise.
For either stopping condition, the Researcher constructs the final context from the accumulated trajectory:
$c^{*} = \operatorname{Finalize}_{\theta}(q, h_T)$, retaining request-relevant information and source provenance for the client agent. If the budget is exhausted without a positive sufficiency assessment, the same finalization step summarizes the evidence collected so far for best-effort answering, without further exploration.

\begin{figure}[t]
    \centering
    \includegraphics[width=0.95\textwidth]{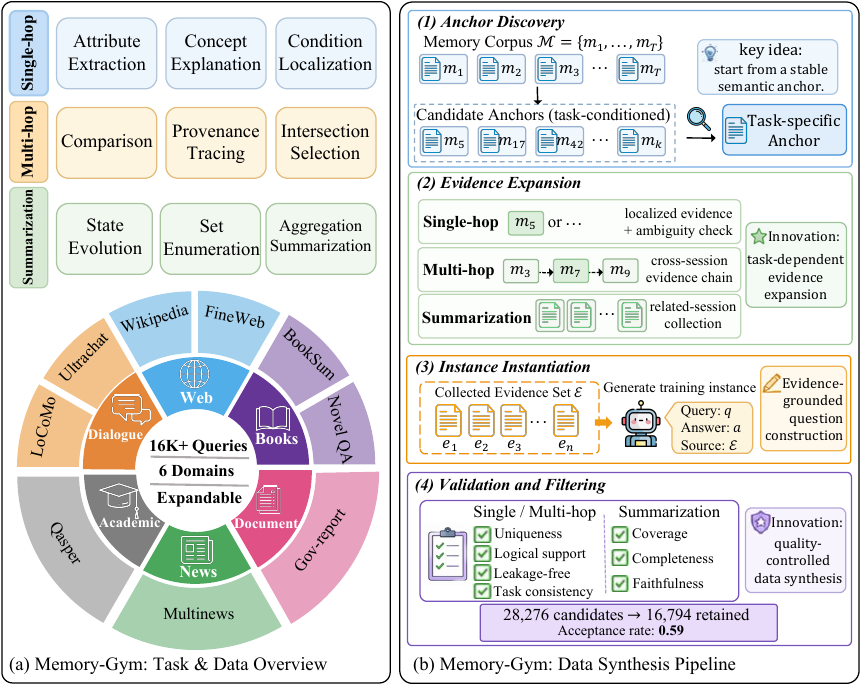} 
    \caption{Overview of Memory-Gym. (a) Memory-Gym organizes realistic memory use into three task families covering nine task types across six heterogeneous domains. (b) Memory-Gym constructs evidence-grounded training instances through a four-stage pipeline.} 
    \label{fig:intro_memorygym} 
\end{figure}

\section{Training JAM with Memory-Gym}
\label{sec:jam_training}

To train the Researcher, we introduce Memory-Gym for evidence-grounded
supervision and a cascaded optimization procedure combining
verified-trajectory SFT with Hint-guided GRPO.

\subsection{Memory-Gym: Data Synthesis for Memory Agents}
\label{subsec:memorygym}

Effectively training the Researcher requires suitable data that captures diverse memory-use scenarios over accumulated histories, which existing QA-style or benchmark-specific training sources often fail to provide. 
To address this data scarcity, we introduce \textbf{Memory-Gym}, a scalable data synthesis pipeline designed for JAM's optimization, as shown in Figure~\ref{fig:intro_memorygym}.

Given a memory corpus $\mathcal{M}=\{s_1,\ldots,s_T\}$ consisting of accumulated sessions and a target task type $z$, Memory-Gym constructs an instance $x=(q,a,\mathcal{E},z)$. 
Here, $q$ is a synthesized request, $a$ is the reference answer, and $\mathcal{E}\subseteq\mathcal{M}$ records the supporting sessions used for construction and validation. 
Below, we introduce the task taxonomy, source domains, and construction pipeline.

\subsubsection{Task Taxonomy and Source Domains}

Memory-Gym organizes realistic memory use over accumulated histories into three task families: single-hop, multi-hop, and multi-session. 
These families focus on the memory operations required by different requests: locating a specific session-level fact, connecting information across sessions, and aggregating observations over broader histories.

\noindent \textbf{Single-hop localization.}
Single-hop tasks require the agent to identify a memory session that directly supports the answer. 
Although the required evidence is local once found, the challenge lies in locating it from a large memory corpus containing irrelevant or semantically similar sessions. 
This family includes attribute extraction, concept explanation, and condition localization.

\noindent \textbf{Multi-hop evidence connection.}
Multi-hop tasks require the agent to combine partial information from multiple sessions. 
The answer cannot be obtained from any single session alone, but must be derived by linking entities, constraints, events, or temporal relations across memory.  
This family includes comparison, provenance tracing, and intersection selection.

\noindent \textbf{Multi-session summarization.}
Summarization tasks require the agent to synthesize a structured answer from a broader collection of related sessions. 
The target answer is often an abstraction over multiple observations, such as temporal changes, repeated mentions, or partially overlapping information. 
This family includes state evolution, set enumeration, and aggregation summarization.

To improve domain coverage, we instantiate these task families over six heterogeneous domains, including web, books, dialogue, news, academic papers, and long-form documents. 
Together, these domains expose memory agents to diverse accumulated histories and provide broad training scenarios for learning realistic memory use.
Detailed data sources are provided in Appendix~\ref{app:data_source}.

\subsubsection{Task Construction Pipeline}

Memory-Gym constructs each training instance through a four-stage pipeline: anchor discovery, task-dependent session expansion, instance instantiation, and validation. 
The pipeline first builds the session set required by the target task type, and then generates a query-answer pair.

\noindent \textbf{Stage 1: Anchor discovery.}
Memory-Gym first selects a task-specific anchor, such as an entity, event, concept, condition, or evolving state, as the semantic starting point of an instance. 
For each sampled source session, it extracts multiple candidate anchors and filters out those that are underspecified, weakly grounded, or unsuitable for the target task type.

\noindent \textbf{Stage 2: Task-dependent session expansion.}
Starting from the anchor, Memory-Gym expands a supporting session set $\mathcal{E}$ according to the memory-use pattern required by the target task family. 
For single-hop tasks, it identifies the local session that directly supports the instance and performs corpus-level ambiguity checking. 
For multi-hop tasks, it retrieves and links related sessions through shared entities, events, constraints, or temporal relations. 
For summarization tasks, it collects a broader set of sessions associated with the same theme, state, or evolving process, so that the instance requires aggregation rather than isolated fact retrieval.

\noindent \textbf{Stage 3: Instance instantiation.}
Given the expanded session set $\mathcal{E}$, Memory-Gym generates a query-answer pair $(q,a)$ consistent with the target task type $z$. 
The query is constructed to require the target memory-use pattern, while the reference answer is derived only from the selected sessions. 
The provenance of supporting sessions is also recorded for validation and downstream training.

\noindent \textbf{Stage 4: Validation and filtering.}
Memory-Gym filters candidates using task-specific quality checks.
For single- and multi-hop tasks, we verify answer uniqueness, evidence support, and the absence of answer leakage; for summarization, we additionally assess coverage, completeness, and faithfulness to the source sessions.
Of \(28{,}276\) candidates, \(16{,}794\) passed validation (\(59.4\%\)).
A manual audit of 200 retained instances found that \(95.0\%\) met all applicable quality criteria.
See Appendices~\ref{app:construction_pipeline} and~\ref{app:memory_gym_human_validation} for construction and human-validation details, respectively.


\subsection{Cascaded Optimization of JAM}

JAM exposes runtime agentic research as a sequential and trainable process. 
We therefore focus on optimizing the Researcher, while keeping the Memorizer fixed to provide a stable hierarchical workspace. 
As shown in Figure~\ref{fig:main}, Memory-Gym supports a cascaded optimization workflow: verified-trajectory SFT initializes the Researcher's tool-use behavior, and Hint-guided GRPO further improves context construction under sparse task-level feedback.

\noindent \textbf{Verified-trajectory SFT.}
To initialize the Researcher with effective thinking, tool-use, and sufficiency-assessment behaviors, we first construct a supervised corpus of verified Researcher trajectories. 
Given a Memory-Gym instance and the workspace constructed by the Memorizer, a strong teacher model solves the task under the JAM workflow and produces an interaction trajectory $\tau$.

We retain trajectories whose final answers are correct and whose exploration processes are judged to be high-quality, using an LLM-as-a-judge to filter unsupported or shortcut behaviors such as shallow search without sufficient inspection. 
The remaining trajectories form a supervised dataset $\mathcal{D}_{\mathrm{SFT}}$.
We fine-tune the Researcher policy by maximizing the likelihood of demonstrated action sets:
\begin{equation}
\mathcal{L}_{\mathrm{SFT}}(\theta)
=
-\sum_{\tau \in \mathcal{D}_{\mathrm{SFT}}}
\sum_{i=1}^{|\tau|}
\log \pi_{\theta}(A_i \mid q,h_{i-1}).
\end{equation}
This stage provides a stable behavioral initialization before policy optimization.

\noindent \textbf{Hint-guided GRPO.}
SFT provides a useful initialization, but multi-step exploration remains unstable under sparse task-level feedback. 
We therefore apply Hint-guided GRPO from the SFT checkpoint, injecting lightweight hints during training rollouts to guide search and tool-use decisions.

Specifically, every $K_h$ steps, a hint generator produces a task-specific hint $\eta_k$ conditioned on the current interaction trajectory and the reference answer:
$\tilde{s}_k = s_k \oplus \eta_k, \text{if } k \bmod K_h = 0.$
The hint provides high-level guidance about missing information, unexplored search directions, or potentially useful tool choices. 
It does not directly reveal the final answer or prescribe a complete search path. 
Hints are used only during training rollouts and are removed at inference time.

We use source recall as the primary reward, since this stage aims to improve the Researcher's ability to locate task-relevant information rather than merely imitate teacher trajectories. 
For a sampled trajectory $\tau$, let $\mathcal{E}_{\tau}$ denote the set of source sessions collected or cited by the Researcher, and let $\mathcal{E}^{*}_{q}$ denote the gold supporting session set provided by Memory-Gym. 
The reward is defined as
$R(\tau)
=
\mathrm{Recall}(\mathcal{E}_{\tau}, \mathcal{E}^{*}_{q})
=
\frac{|\mathcal{E}_{\tau}\cap \mathcal{E}^{*}_{q}|}
{|\mathcal{E}^{*}_{q}|}.$
For each query, we sample a group of $G$ trajectories $\{\tau_j\}_{j=1}^{G}$ and compute their rewards $\{R(\tau_j)\}_{j=1}^{G}$. 
The advantage $A_j$ is obtained by normalizing rewards within the group. 
We then optimize the Researcher with the GRPO objective:
\begin{equation}
J_{\mathrm{GRPO}}(\theta)
=
\mathbb{E}\!\left[
\frac{1}{G}\sum_{j=1}^{G}
\min\!\left(
\rho_j A_j,\,
\mathrm{clip}(\rho_j,1-\epsilon,1+\epsilon)A_j
\right)
-\beta D_{\mathrm{KL}}(\pi_{\theta}\|\pi_{\mathrm{ref}})
\right],
\end{equation}
where $\rho_j$ denotes the policy ratio for $\tau_j$, $\pi_{\mathrm{ref}}$ is the reference policy, and $\beta$ controls the KL penalty.

\section{Experiments}
\label{sec:experiment}

We conduct extensive experiments to evaluate JAM and validate the effectiveness of Memory-Gym as a training source. 
Specifically, our evaluation aims to answer the following research questions:
\textbf{RQ1}: How does JAM perform compared with existing AOT-style memory systems and trained memory agents?
\textbf{RQ2}: Does Memory-Gym provide effective and transferable supervision?
\textbf{RQ3}: How do the key components of JAM contribute to the overall performance?
\textbf{RQ4}: How does JAM benefit from increased test-time computation, and what is its effectiveness--efficiency trade-off?

\subsection{Experiment Setting}

\noindent \textbf{Datasets \& Metrics.}
We evaluate JAM on four long-context and memory-intensive benchmarks: LoCoMo~\cite{locomo}, LongMemEval (LME)~\cite{wu2024longmemeval}, NarrativeQA (NAQA)~\cite{kovcisky2018narrativeqa}, and HotpotQA~\cite{yu2025memagent}. 
These benchmarks cover long-term conversational memory, interactive memory, long-document reasoning, and multi-hop reasoning over dispersed evidence. 
We use the official evaluation metric for each benchmark: F1 for LoCoMo, NarrativeQA, and HotpotQA, and accuracy for LongMemEval.

\noindent \textbf{Baselines.} 
We compare JAM against both training-free and training-based baselines.
Training-free baselines include Vanilla LLM and RAG~\cite{jiang2023active}, as well as AOT-style memory systems A-MEM~\cite{xu2025mem}, Mem0~\cite{chhikara2025mem0}, MemoryOS~\cite{kang2025memory}, and LightMem~\cite{fang2025lightmem}.
Training-based baselines include MEM1~\cite{zhou2026mem1}, MemAgent~\cite{yu2025memagent}, and Memory-R1~\cite{yan2025memory}.

\noindent \textbf{Implementation Details.}
We use Qwen3.5-4B~\cite{qwen3.5} as the backbone model for JAM and all AOT-style baselines.
For training-based baselines, we use their released checkpoints or official reported settings.
JAM uses BM25 and BGE-M3~\cite{chen2024bge} as its default lexical and dense retrievers.
Further implementation and training details are provided in Appendix~\ref{app:training_implementation}.



\begin{table*}[t]
\centering
\setlength{\tabcolsep}{3pt}
\caption{Results of JAM and baselines on benchmarks. Best results are in \textbf{bold}; second-best results are \underline{underlined}. Single-hop(SH); Multi-hop(MH); Temporal(TE); Open Domain(OD); Overall(OA). \\$^\dagger$: As Memory-R1 is not publicly released, we faithfully reported the results from the original paper.}
\resizebox{\linewidth}{!}{%
\begin{tabular}{llcccccccccc}
\toprule
\multirow{2}{*}{\textbf{Method}} 
& \multirow{2}{*}{\textbf{Base LLM}} 
& \multicolumn{5}{c}{\textbf{LoCoMo}} 
& \multirow{2}{*}{\textbf{LME}} 
& \multirow{2}{*}{\textbf{NAQA}} 
& \multicolumn{3}{c}{\textbf{HotpotQA}} \\
\cmidrule(lr){3-7} \cmidrule(lr){10-12}
& 
& \textbf{SH} 
& \textbf{MH} 
& \textbf{TE} 
& \textbf{OD} 
& \textbf{OA}
& 
& 
& \textbf{56k} 
& \textbf{112k} 
& \textbf{224k} \\
\midrule
\textbf{\textit{Training-free}} &  &  &  &  &  &  &  &  &  &  & \\
VANILLA        & Qwen3.5-4B   & 48.15 & 32.98 & 42.66 & 19.63 & 42.45 & 54.80 & 31.23 & \underline{63.56} & \underline{53.04} & 36.97 \\
RAG            & Qwen3.5-4B   & 49.04 & 34.39 & 44.11 & 17.68 & 43.37 & 56.20 & \underline{32.86} & 49.97 & 44.83 & \underline{49.87} \\
A-MEM          & Qwen3.5-4B   & 46.99 & 29.88 & 42.56 & 18.65 & 41.17 & \underline{56.80} & 31.30 & 27.08 & 25.39 & 28.32 \\
Mem0           & Qwen3.5-4B   & 34.84 & 26.45 & 34.47 & 16.83 & 32.10 & 34.40 & 26.72 & 30.19 & 27.57 & 24.53 \\
MemoryOS       & Qwen3.5-4B   & 32.75 & 28.95 & 36.36 & 14.50 & 31.67 & 36.00 & 27.21 & 20.27 & 22.49 & 19.61 \\
LightMem       & Qwen3.5-4B   & 35.35 & 27.88 & 35.47 & 18.68 & 32.97 & 52.40 & 28.24 & 30.73 & 27.12 & 26.51 \\
\cmidrule(lr){1-12}
\textbf{\textit{Trained}}     &  &  &  &  &  &  &  &  &  &  & \\
MEM1           & MEM1-7B      & 28.56 & 20.48 & 32.46 & 14.73 & 27.11 & 32.40 & 23.86 & 33.94 & 32.47 & 27.32 \\
MemAgent       & MemAgent-14B & \underline{49.19} & \underline{36.33} & \underline{53.77} & 22.50 & \underline{46.13} & 55.20 & 28.24 & 56.26 & 50.70 & 46.88 \\
Memory-R1-PPO$^\dagger$   & Mem-R1-8B$^\dagger$    & 32.52 & 26.86 & 41.57 & \underline{45.30} & 41.05 & --    & --    & --    & --    & --    \\
Memory-R1-GRPO$^\dagger$  & Mem-R1-8B$^\dagger$    & 35.73 & 35.65 & 49.86 & \textbf{47.42} & 45.02 & --    & --    & --    & --    & --    \\
\cmidrule(lr){1-12}
\textbf{\textit{Our Method JAM}} &  &  &  &  &  &  &  &  &  &  & \\
with training & Qwen3.5-4B    & \textbf{55.14} & \textbf{40.21} & \textbf{62.59} & 25.12 & \textbf{52.09} & \textbf{65.20} & \textbf{45.72} & \textbf{64.56} & \textbf{59.28} & \textbf{58.69} \\
\bottomrule
\end{tabular}%
}
\vspace{-3mm}
\label{tab:main}
\end{table*}

\subsection{Main Results}

Table~\ref{tab:main} reports the performance of JAM and ten competitive baselines across four benchmarks, demonstrating the overall effectiveness of our framework for memory-intensive tasks.

First, JAM exhibits strong generalization across heterogeneous settings. While several AOT-style memory systems remain competitive on dialogue-centric benchmarks such as LoCoMo, their performance often degrades on long-document and multi-hop reasoning benchmarks. In contrast, JAM remains consistently strong across different task types, suggesting that its hierarchical workspace and runtime agentic research enable more robust memory use across domains.


Second, JAM performs strongly even against substantially larger trained memory agents. Despite using a Qwen3.5-4B backbone, JAM outperforms 7B--14B trained baselines across most settings. The results highlight the importance of effective training supervision: Memory-Gym exposes the Researcher to diverse memory-use scenarios, enabling it to learn exploration behaviors that generalize across heterogeneous benchmarks. We examine the contribution of Memory-Gym supervision more directly in Section~\ref{sec:memory_gym_supervision}.

Third, JAM remains stable as context length increases. On HotpotQA with increasing context sizes, prior methods often degrade when relevant information is buried among longer distractor contexts, while JAM sustains comparatively strong performance. This suggests that JAM can better handle dispersed information because its offline workspace organization and trained Researcher support progressive, query-conditioned exploration at runtime.

Complementary analyses further show that these gains are stable across inference seeds and persist under an LLM-judge evaluation (Appendices~\ref{app:inference_variance} and~\ref{app:llm_judge}).

\subsection{Effectiveness and Transferability of Memory-Gym}
\label{sec:memory_gym_supervision}

\begin{wraptable}{r}{0.52\linewidth}
\vspace{-\intextsep}
\centering
\small
\setlength{\tabcolsep}{2.3pt}
\caption{Memory-Gym supervision analysis. All trained variants use SFT only.}
\renewcommand{\arraystretch}{1.02}
\begin{tabular}{@{}lrrrr@{}}
\toprule
Training data & LoCoMo & LME & NAQA & HotpotQA \\
\midrule
\textit{w/o} SFT
    & 48.59 & 48.80 & 34.98 & 45.44 \\
\midrule
MemAgent data
    & 48.60 & 55.20 & 31.98 & \textbf{54.53} \\
Memory-R1 data
    & 48.90 & 57.20 & 32.42 & 49.81 \\
\midrule
SH only
    & 50.59 & 58.80 & 34.40 & 50.45 \\
MH only
    & 48.97 & 57.60 & 37.42 & 51.96 \\
SU only
    & 49.43 & 59.80 & 40.94 & 50.86 \\
\midrule
Memory-Gym
    & \textbf{50.70}
    & \textbf{60.20}
    & \textbf{41.59}
    & 52.08 \\
\bottomrule
\end{tabular}
\label{tab:memory_gym_supervision}
\end{wraptable}

To evaluate Memory-Gym, we first compare different supervision sources
and task families under a fixed JAM framework and backbone, with all
trained variants using SFT only. We then examine whether the learned
memory-use behaviors transfer beyond Memory-Gym's source domains on
LongCodeQA~\cite{rando2025longcodebench}.

\noindent\textbf{Supervision sources and task diversity.}
Table~\ref{tab:memory_gym_supervision} shows that alternative supervision
sources yield uneven gains across benchmarks. MemAgent-derived supervision
performs best on HotpotQA, whereas both MemAgent- and Memory-R1-derived
supervision underperform the untrained Researcher on NarrativeQA.
In contrast, Memory-Gym improves over the untrained Researcher on all
four benchmarks and achieves the strongest source-comparison results on
three. The task-family analysis shows a complementary pattern: among
single-family variants, SH performs best on LoCoMo, MH on HotpotQA, and
SU on LongMemEval and NarrativeQA, while the full mixture
outperforms every single-family variant across all four benchmarks.
These results are consistent with complementary benefits from evidence
localization, cross-session connection, and broader aggregation, motivating
the use of diverse memory-use tasks for supervision.

\begin{wraptable}{r}{0.40\linewidth}
\vspace{-\intextsep}
\centering
\small
\setlength{\tabcolsep}{2.6pt}
\caption{Cross-domain transfer on LongCodeQA.}
\renewcommand{\arraystretch}{1.02}
\begin{tabular}{@{}lrrrr@{}}
\toprule
Researcher & 64K & 128K & 256K & Overall \\
\midrule
Untrained & 56.58 & 55.43 & 49.23 & 54.08 \\
SFT only  & 68.42 & 64.13 & 64.62 & 65.67 \\
SFT+RL    & \textbf{76.32}
          & \textbf{76.09}
          & \textbf{73.85}
          & \textbf{75.54} \\
\bottomrule
\end{tabular}
\label{tab:longcodeqa_transfer}
\end{wraptable}

\noindent\textbf{Cross-domain transfer.}
We further evaluate the Researcher on LongCodeQA, a code-comprehension
benchmark from a domain not represented in Memory-Gym, without using any
LongCodeQA examples for training or adaptation. As shown in
Table~\ref{tab:longcodeqa_transfer}, overall accuracy increases from
54.08\% for the untrained Researcher to 65.67\% after SFT and 75.54\%
after the full SFT+RL pipeline. Improvements are consistent across the
64K, 128K, and 256K subsets. These results indicate that the memory-use
behaviors learned from Memory-Gym transfer beyond its source domains,
with policy optimization providing further gains.

\subsection{Ablation Study}

\begin{wraptable}{r}{0.52\textwidth}
\vspace{-\intextsep}
\vspace{-8pt}
\centering
\small
\setlength{\tabcolsep}{2.3pt}

\caption{Ablation results on different benchmarks. LoCoMo denotes the overall score. HotpotQA denotes the average over 56k, 112k, and 224k settings.}
\label{tab:ablation}

\begin{tabular}{lcccc}
\toprule
\textbf{Method} 
& \textbf{LoCoMo} 
& \textbf{LME} 
& \textbf{NAQA} 
& \textbf{HotpotQA} \\
\midrule
\textit{w/o} Training    & 48.59 & 48.80 & 34.98 & 45.44 \\
SFT Only                 & 50.70 & 60.20 & 41.59 & 52.08 \\
GRPO \textit{w/o} Hint   & 51.42 & 62.40 & 43.62 & 55.64 \\
\textit{w/o} Memorizer   & 49.06 & 64.00 & 38.87 & 57.72 \\
\textit{w/o} Researcher  & 43.30 & 58.40 & 32.51 & 39.10 \\
\textit{w/o} BM25        & 45.90 & 62.40 & 38.88 & 52.14 \\
\textit{w/o} Embedding   & 50.86 & 61.80 & 39.45 & 58.13 \\
Full                     & 52.09 & 65.20 & 45.72 & 60.84 \\
\bottomrule
\end{tabular}

\vspace{-2mm}
\end{wraptable}
To evaluate the contribution of each core component, we conduct systematic ablation studies by removing individual modules. The results are summarized in Table~\ref{tab:ablation}. 
First, Researcher optimization consistently improves performance. 
The untrained variant already benefits from the JAM framework, but remains limited on more challenging memory-intensive tasks. SFT improves the Researcher by providing high-quality trajectory supervision for thinking, tool use, and workspace exploration. GRPO without hints brings further gains over SFT-only, showing that reinforcement learning helps the Researcher improve source discovery beyond imitation. Adding hint guidance achieves the best performance, especially on NAQA and HotpotQA, indicating that lightweight hints stabilize multi-step exploration under sparse rewards.
Second, both the Memorizer and the Researcher contribute substantially
to JAM. Removing the Memorizer weakens performance by eliminating the
hierarchical organization and navigational summaries of the workspace.
Removing the Researcher causes an even larger degradation, highlighting
the importance of iterative, query-conditioned exploration at runtime.
A controlled comparison against a flat raw-history store further shows
that hierarchical workspace organization improves both effectiveness and
online efficiency: it improves LoCoMo F1 from 49.06 to 52.09 while
reducing online latency from 17.11 to 13.81 seconds per query and average
research rounds from 3.57 to 3.16
(Appendix~\ref{app:efficiency_details}).
We further examine the sensitivity to the Memorizer backbone in
Appendix~\ref{app:memorizer_backbone}, where scaling to larger backbone
models yields no consistent downstream improvement.
Finally, removing either BM25 or dense retrieval also degrades performance, indicating that lexical and semantic retrieval are complementary and jointly provide more informative candidates for downstream exploration.

\subsection{Test-Time Scalability and Efficiency}
\label{sec:scalability_efficiency}

\begin{figure}[t]
    \centering
    \begin{minipage}[t]{0.6\textwidth}
        \vspace{-3mm}
        \centering
        \includegraphics[width=\linewidth]{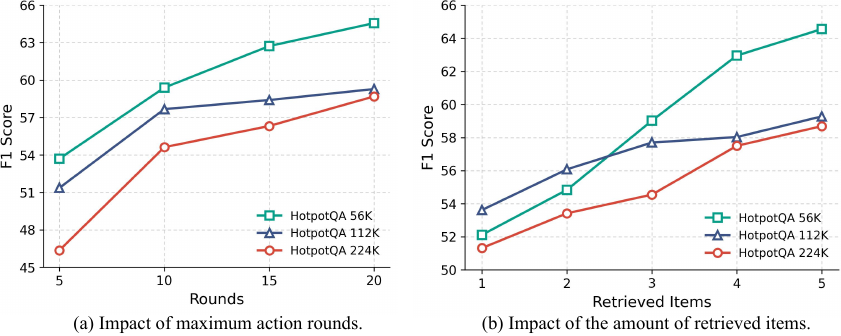}
        \caption{
        Test-time scaling with increasing action rounds
        (left) and retrieved items (right).
        }
        \label{fig:scalability}
    \end{minipage}
    \hfill
    \begin{minipage}[t]{0.38\textwidth}
        \vspace{0pt}
        \centering
        \includegraphics[width=\linewidth]{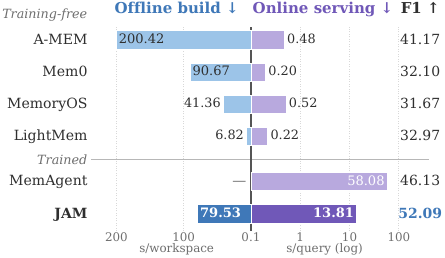}
        \caption{
        Wall-clock latency and overall F1 on LoCoMo.
        }
        \label{fig:wallclock_latency}
    \end{minipage}
    \vspace{-2mm}
\end{figure}

\noindent\textbf{Scaling test-time computation.}
Figure~\ref{fig:scalability} examines two forms of inference-time budget:
the maximum number of action rounds and the number of retrieved items per
search. Increasing the action budget from 5 to 20 consistently improves
performance, providing the Researcher with more opportunities to refine
its search directions and inspect candidate workspace regions, with
diminishing gains at larger budgets. Increasing the number of retrieved
items from 1 to 5 shows a similar trend by broadening the candidate set
available at each search step.

Importantly, the maximum action budget provides exploration headroom
rather than imposing a fixed inference cost. Under the default 20-round
limit, the median number of rounds is only 3 on LoCoMo, LongMemEval, and
NarrativeQA and 6 on HotpotQA, while budget exhaustion remains below
5\% across all benchmarks. Thus, JAM varies its computation across queries
and terminates well before the maximum budget in most cases. Detailed
runtime statistics are in Appendix~\ref{app:runtime_behavior}.

\noindent\textbf{Runtime efficiency.}
Figure~\ref{fig:wallclock_latency} reports wall-clock latency on LoCoMo,
separating one-time offline workspace construction from online query
serving. JAM requires 79.53\,s to construct a workspace, lower than A-MEM
(200.42\,s) and Mem0 (90.67\,s), but higher than MemoryOS (41.36\,s) and
LightMem (6.82\,s). This cost is incurred once, after which the resulting
workspace is reused across queries.
The iterative Researcher introduces additional query-time latency relative
to one-shot memory systems. However, this additional computation is
accompanied by substantially higher answer quality: JAM achieves an
overall LoCoMo F1 of 52.09, outperforming all AOT-style baselines despite
their lower serving latency. Compared with the trained agentic baseline
MemAgent, JAM reduces query latency from 58.08\,s to 13.81\,s (76.2\%)
while improving F1 from 46.13 to 52.09. Together, these results indicate
a favorable quality--latency trade-off: JAM spends more computation than
lightweight one-shot memory systems to obtain markedly higher answer
quality, while remaining substantially faster than MemAgent.
Complementary token-consumption results show a similar quality--cost
pattern, with JAM achieving the highest F1 while using substantially
fewer tokens per task than MemAgent
(Appendix~\ref{app:efficiency_details}).


\section{Related Work}

\subsection{AOT-style Agent Memory Systems}

Recent studies on agent memory systems mainly focus on preserving and organizing accumulated contexts for AI agents~\cite{packer2023memgpt}. 
A large body of work follows an AOT design philosophy, where historical observations are compressed or structured into pre-constructed memories before specific online requests arrive. 
Representative methods organize observations into linked memory notes, hierarchical memory stores, or lightweight summarization and consolidation pipelines~\cite{xu2025mem,kang2025memory,fang2025lightmem}. 
Nevertheless, their request-agnostic preprocessing can discard fine-grained details or cross-session dependencies that later become important. 
Moreover, stored memories are often utilized through relatively simple retrieval or summarization procedures, which are less effective for complex memory-intensive tasks that require iterative inspection and query-conditioned context construction.

\subsection{Training and Data for Memory Agents}

Recent work has trained memory agents with SFT or RL to improve memory use in long-context scenarios~\cite{wang2025mem,shi2025look}. 
These studies demonstrate the promise of optimization-based memory agents, but their supervision is often tied to constrained multi-hop QA datasets such as HotpotQA~\cite{yu2025memagent} or benchmark-specific splits~\cite{yu2025memagent,yan2025memory,yue2026mem}. 
This limits the diversity of memory-use scenarios and makes it difficult to train agents for realistic settings that require localization, cross-session connection, temporal tracking, and aggregation over accumulated long contexts.

\section{Conclusion}

In this paper, we propose JAM, a Just-In-Time agent memory framework that
shifts memory from request-agnostic pre-construction to runtime
query-conditioned context construction.
JAM combines a Memorizer that preserves and organizes accumulated
histories in a hierarchical workspace with a Researcher that iteratively
explores the workspace to construct task-relevant context.
To optimize this process, we introduce Memory-Gym and a cascaded training
workflow combining verified-trajectory SFT with Hint-guided GRPO.
Experiments across diverse long-context and memory-intensive benchmarks
show that JAM consistently improves task performance over AOT-style memory
systems, transfers across domains, and offers a favorable quality--cost
trade-off for runtime memory construction.

\bibliography{iclr2027_conference}
\bibliographystyle{iclr2027_conference}

\clearpage

\appendix

\section{Details of Memory-Gym}
\label{app:Memory-Gym_details}
\subsection{Data Sources and Corpus Construction}
\label{app:data_source}
Memory-Gym is built from six heterogeneous domains: web, books, dialogue, news, academic papers, and long-form documents. These domains are selected to cover diverse forms of accumulated long-context memory, including factual lookup, narrative reasoning, conversational memory, event evolution, scholarly evidence tracing, and structured document understanding. We normalize all raw inputs into memory corpora $\mathcal{M}=\{s_1,\ldots,s_T\}$, where each $s_i$ denotes a memory session. The following paragraphs describe the domain-specific data sources and session construction procedures.

\noindent \textbf{Web.}
The web domain is constructed from Wikipedia~\cite{wikimedia_wikipedia_2023} and FineWeb~\cite{penedo2024fineweb}, covering both structured encyclopedic knowledge and diverse open-domain web content. 
FineWeb provides cleaned general web documents with broad topics and heterogeneous writing styles. 
We retain FineWeb documents whose token count falls within $[30{,}000, 150{,}000]$ and process each retained document with the JAM Memorizer to obtain memory sessions. 
Wikipedia provides relatively clean, structured, and entity-centric encyclopedic articles. 
Since individual articles are usually short, we construct long-context memory corpora through random composition: each session is formed by concatenating 5--8 randomly sampled articles, and each memory corpus contains 30--50 such sessions. 
This creates web memory corpora with many factual units and distractor articles for evidence localization across sessions.

\noindent \textbf{Books.}
The books domain is constructed from BookSum~\cite{kryscinski2022booksum} and NovelQA~\cite{novelqa}, both of which provide long-form narrative texts involving extended plots, characters, events, and state changes. 
BookSumBooks contains book-length literary texts, while NovelQA consists of full-length novels designed for long-context narrative understanding.
For both sources, we filter candidate books by length and retain full texts whose token count falls within $[20{,}000, 500{,}000]$. 
Each retained book is treated as a long-form raw input and processed by the JAM Memorizer, which segments it into memory sessions and organizes them into the hierarchical workspace. 
This preserves the original narrative order while converting each book into a session-based memory corpus.

\noindent \textbf{Dialogue.}
The dialogue domain is constructed from UltraChat~\cite{ding2023enhancing} and LoCoMo~\cite{locomo}, which provide multi-turn conversational data for simulating accumulated conversational contexts. 
UltraChat is a large-scale open-domain dialogue dataset covering diverse user intents, topics, and conversational styles. 
For UltraChat, we retain conversations with at least 5 turns and treat each retained conversation as one memory session. 
For each memory corpus, we randomly sample 25--50 such sessions, resulting in an accumulated dialogue memory corpus composed of multiple conversational episodes.
LoCoMo is a long-term conversational memory benchmark designed to evaluate reasoning over information accumulated across extended interactions. 
We use the official scripts to generate dialogue data. 
Each generated session contains more than 15 dialogue turns, and each memory corpus contains more than 30 sessions. 
Compared with UltraChat, LoCoMo provides longer and more memory-intensive conversational contexts for cross-session evidence localization and temporal tracking.

\noindent \textbf{News.}
The news domain is constructed from MultiNews~\cite{fabbri2019multi}, which contains articles about diverse real-world events, topics, and social issues. 
Since related news articles often describe evolving events from different perspectives, this domain supports tasks involving event tracking, provenance tracing, comparison, and aggregation. 
We retain articles with at least 1,000 tokens, encode them using BGE-M3, and cluster them based on cosine similarity. 
For each memory corpus, we randomly sample 30--35 articles from a target cluster, with each article treated as one memory session. 
This yields long-context news memory corpora with distributed evidence and realistic distractor reports.

\noindent \textbf{Academic.}
The academic domain is constructed from Qasper~\cite{dasigi2021dataset}, a question-answering dataset grounded in academic papers. 
Since academic papers contain structured information about research problems, methods, experiments, and results, this domain supports tasks such as concept explanation, method comparison, and evidence tracing. 
We use the original section structure of each paper for corpus construction, treating each section as one memory session. 
Papers with fewer than 10 sections are filtered out to ensure sufficient cross-session structure for long-context reasoning.

\noindent \textbf{Document.}
The document domain is constructed from Gov-report~\cite{huang2021efficient}, which contains long-form government reports with hierarchical section structures. 
These reports describe policies, programs, institutional plans, outcomes, and conditions, making them suitable for condition localization, state tracking, set enumeration, and aggregation tasks. 
For corpus construction, we preserve the original document hierarchy by recursively traversing the report tree. 
Each node is treated as one memory session, and its child nodes are further expanded into additional sessions.
We retain reports with at least 28 recursively expanded sessions to ensure sufficient depth and breadth for evidence localization and reasoning over structured long documents.

\noindent \textbf{Evaluation contamination control.}
To avoid evaluation contamination, we ensure that all memory corpora used for Memory-Gym construction are disjoint from the evaluation data used in our experiments. We exclude all released evaluation instances, including their questions, answers, supporting evidence, and source sessions, from Memory-Gym construction. In particular, although LoCoMo is included as a dialogue-domain source, we do not use any released LoCoMo evaluation conversations, questions, answers, or supporting evidence. Instead, we generate new LoCoMo-style dialogue corpora using the official data-generation pipeline, with independently sampled scenarios and sessions. Therefore, the Memory-Gym training corpora have no overlap with the evaluation benchmarks in terms of users, conversations, sessions, queries, answers, or supporting evidence.

\subsection{Implementation Details of the Construction Pipeline}

We use MiniMax-2.5~\cite{minimax2026m25} as the LLM-based generator for anchor extraction and evidence-grounded question-answer instantiation. Given a memory corpus and a target task type, the generator identifies candidate anchors from the available sessions and constructs query-answer pairs grounded in the selected evidence. We also use MiniMax-2.5 as the judge model to validate generated instances according to task-specific criteria. We use separate prompts for generation and validation to reduce the chance that the same generation procedure accepts invalid instances.
For retrieval and clustering, we use BGE-M3~\cite{chen2024bge} as the embedding model. Candidate documents or sessions are encoded into dense vectors, and cosine similarity is used to measure semantic relatedness. The resulting similarity scores are used for clustering related documents during corpus construction and for retrieving candidate evidence sessions during task construction.

\subsection{Task-specific Construction and Validation}

\label{app:construction_pipeline}
Although Memory-Gym follows a unified construction pipeline, different task types require different evidence structures and validation criteria. We therefore define task-specific rules to ensure that each generated query reflects the intended memory-use pattern, including single-session localization, cross-session reasoning, and multi-session aggregation. Across all tasks, valid instances must be evidence-supported, leakage-free, and consistent with the target task type.

\noindent \textbf{Single-hop tasks.}
Single-hop tasks require the answer to be supported by a single memory session, while the challenge is to locate the correct session from a large memory corpus. We construct three single-hop task types: attribute extraction, concept explanation, and condition localization.

\textit{Attribute Extraction.}
For attribute extraction, we randomly sample a memory session and ask the generator to identify salient attributes of entities, objects, events, or concepts in the session. We then generate a query about one selected attribute, with the attribute value used as the reference answer. For validation, we retrieve passages using the generated query and discard instances if alternative plausible answers are found. The judge model further checks whether the query-answer pair is reasonable, evidence-supported, and consistent with the attribute extraction task.

\textit{Concept Explanation.}
For concept explanation, we randomly sample a session and ask the generator to identify key concepts or objects that have explicit definitions or property descriptions. The query asks for the definition or properties of the selected concept, and the answer is derived from the corresponding description. During validation, we retrieve relevant passages to check whether other sessions contain competing definitions or descriptions, and further verify that the concept is uniquely grounded in the target session. The judge model then checks evidence support and task consistency.

\textit{Condition Localization.}
For condition localization, we randomly sample a session and ask the generator to identify causal, conditional, or prerequisite relations. We extract the resulting event or phenomenon as the query target and generate a question asking what condition or cause leads to it. The answer is the corresponding condition or cause in the evidence session. For validation, the judge model checks whether the answer has a valid causal or prerequisite relation with the queried outcome, and we discard instances if other sessions provide conflicting causes or conditions.

\paragraph{Multi-hop tasks.}
Multi-hop tasks require the answer to be derived by connecting evidence from multiple memory sessions. We construct three types of multi-hop tasks: comparison, provenance tracing, and intersection selection.

\textit{Comparison.}
For comparison, we randomly sample several sessions and ask the generator to identify a feature dimension suitable for entity comparison, such as location, scale, membership, or outcome. We then use this feature as a query to retrieve relevant sessions and ask the generator to identify two to three entities with corresponding evidence. A comparison question is generated based on the selected entities and feature dimension. For validation, the judge checks whether the answer is supported by the retrieved evidence, correctly addresses the comparison, and remains unique after searching each compared entity in the memory corpus.

\textit{Provenance Tracing.}
For provenance tracing, we construct cross-session entity-linking tasks. We first select a target entity that appears in at least two different sessions. One session provides a unique abstract description of the entity without mentioning its name, while another session provides an event or fact associated with the entity. The question is generated using the abstract description and event details, but the entity name is strictly removed. For validation, we apply name-leakage checks, verify that the abstract description uniquely identifies the target entity, and ask the judge to confirm that the full reasoning chain from identity resolution to event grounding is valid.

\textit{Intersection Selection.}
For intersection selection, we construct questions whose target can only be identified by combining multiple weak constraints. Starting from a target entity, we retrieve several sessions containing it and ask the generator to extract broad descriptions from each session. Each individual description should be ambiguous, while their intersection should uniquely identify the target. The generated question combines these constraints and asks about a specific fact or event without revealing the entity name. The judge validates individual ambiguity, intersection-level uniqueness, and answer correctness.

\paragraph{Summarization tasks.}
Summarization tasks require the agent to aggregate information over multiple memory sessions rather than locating a single evidence unit. We construct three types of summarization tasks: state evolution, set enumeration, and aggregation summarization.

\textit{State Evolution.}
For state evolution, we first sample five consecutive sessions and ask the generator to identify entities whose states change over time. After selecting a target entity, we scan the full memory corpus in chronological order by batches of sessions and ask the generator to identify all sessions that describe the entity's state changes. The collected sessions are then used to generate a question about the evolution of the target entity, with the answer summarizing the complete state transition process. For validation, the judge model checks whether the question is reasonable and whether the answer faithfully reflects the state changes supported by the selected sessions.

\textit{Set Enumeration.}
For set enumeration, we first cluster sessions using embeddings and FAISS to identify semantically related contexts. From a sampled cluster, the generator identifies an extraction criterion defined by a category and a constraint, such as entities of a certain type mentioned under a specific condition. We then scan the full memory corpus to incrementally collect all unique entities satisfying the criterion, together with their source sessions. The final query is generated from the extraction criterion, and the answer consists of the collected entity set. For validation, the judge model checks both correctness and completeness, ensuring that all listed items are supported by evidence and that no supported items are omitted.

\textit{Aggregation Summarization.}
For aggregation summarization, we cluster sessions based on embedding similarity and ask the generator to infer a coherent topic from the top sessions in a cluster. We then use this topic to scan the full memory corpus in batches and collect all sessions relevant to the topic. Given the collected evidence, the generator constructs a question asking for an aggregated summary and produces a reference answer. During validation, the judge model checks whether the instance matches the aggregation summarization task, whether the question and answer are aligned, and whether the answer is faithful to the relevant sessions.

\subsection{Human Validation of Memory-Gym}
\label{app:memory_gym_human_validation}

We assess the quality of Memory-Gym instances and the reliability of automatic validation through three complementary audits: human--LLM agreement on candidates before filtering, a human quality audit of retained instances, and an error analysis of rejected candidates.

\paragraph{Annotation protocol.}
Across all three audits, two primary annotators independently apply predefined task-specific criteria while blinded to the automatic decisions and each other's labels. A third annotator adjudicates disagreements. Inter-annotator agreement and Cohen's $\kappa$ are computed from the independent labels before adjudication, whereas quality estimates and human--LLM comparisons use the adjudicated labels.
The quality criteria follow the task-specific validation requirements described in Appendix~A.3. All instances are assessed for evidence support, absence of answer leakage in the query, and task consistency. Single-hop and multi-hop instances are additionally assessed for answer uniqueness, while summarization instances are assessed for coverage, completeness, and faithfulness to the collected sessions. An instance passes the overall quality assessment only if it satisfies all criteria applicable to its task type.

\paragraph{Human--LLM agreement before filtering.}
We sample 200 candidates from the pre-filter pool, stratified across all nine task types and six source domains. The two primary annotators achieve 96.0\% agreement and Cohen's $\kappa=0.916$. After adjudication, 124 candidates are accepted, corresponding to a human acceptance rate of 62.0\%.
We evaluate the same candidates with MiniMax-2.5, GPT-5.5, and Qwen3.5-122B-A10B using identical evidence, validation criteria, prompts, and output formats across the three models. MiniMax-2.5 is the judge used in the construction pipeline; the other two models provide additional comparisons against the human reference. Precision, recall, and F1 treat acceptance as the positive class.
As shown in Table~\ref{tab:mg_human_llm_agreement}, all three judges achieve over 90\% agreement and $\kappa>0.80$ with the adjudicated human labels. Their acceptance rates are slightly below the human reference, suggesting a moderately conservative tendency on this sample.

\begin{table}[t]
\centering
\caption{Human--LLM agreement on 200 pre-filter candidates. The adjudicated human labels serve as the reference. Acceptance is the positive class for precision, recall, and F1. All values except Cohen's $\kappa$ are percentages.}
\label{tab:mg_human_llm_agreement}
\small
\setlength{\tabcolsep}{3.5pt}
\begin{tabular}{@{}lrrrrrr@{}}
\toprule
Evaluator & \shortstack{Accept\\rate} & Agreement & $\kappa$ & Precision & Recall & F1 \\
\midrule
Human reference & 62.0 & -- & -- & -- & -- & -- \\
MiniMax-2.5 & 58.5 & 91.5 & 0.823 & 95.7 & 90.3 & 92.9 \\
GPT-5.5 & 60.0 & 92.0 & 0.832 & 95.0 & 91.9 & 93.4 \\
Qwen3.5-122B-A10B & 57.5 & 90.5 & 0.803 & 95.7 & 88.7 & 92.1 \\
\bottomrule
\end{tabular}
\end{table}

\paragraph{Human audit of retained instances.}
We sample 200 instances from the 16,794 retained Memory-Gym instances, including 134 single-hop or multi-hop instances and 66 summarization instances. Table~\ref{tab:mg_retained_human_audit} reports criterion-level quality and inter-annotator agreement. Overall, 190 of the 200 instances satisfy all applicable criteria, yielding a 95.0\% pass rate with a 95\% Wilson confidence interval of 91.0\%--97.3\%. Criterion-level pass rates range from 95.5\% to 99.5\%. These results support the quality of the retained instances while identifying a small proportion with residual quality issues.

\begin{table}[t]
\centering
\caption{Human audit of 200 retained Memory-Gym instances. Applicable denotes the number of instances assessed under each criterion. Passed counts and pass rates use adjudicated labels; agreement and Cohen's $\kappa$ use the two primary annotators' labels before adjudication.}
\label{tab:mg_retained_human_audit}
\small
\setlength{\tabcolsep}{3.5pt}
\begin{tabular}{@{}lrrrrr@{}}
\toprule
Quality criterion & Applicable & Passed & \shortstack{Pass rate\\(\%)} & \shortstack{Agreement\\(\%)} & $\kappa$ \\
\midrule
Evidence support & 200 & 196 & 98.0 & 99.5 & 0.886 \\
Answer uniqueness & 134 & 131 & 97.8 & 99.3 & 0.853 \\
Leakage-free & 200 & 199 & 99.5 & 100.0 & 1.000 \\
Task consistency & 200 & 197 & 98.5 & 99.5 & 0.855 \\
Coverage & 66 & 64 & 97.0 & 98.5 & 0.792 \\
Completeness & 66 & 63 & 95.5 & 98.5 & 0.849 \\
Faithfulness & 66 & 65 & 98.5 & 100.0 & 1.000 \\
\midrule
Pass all applicable criteria & 200 & 190 & 95.0 & 97.5 & 0.787 \\
\bottomrule
\end{tabular}
\end{table}

\paragraph{Error analysis of rejected candidates.}
We sample 100 rejected candidates across task types and domains. Annotators independently assess candidate validity and assign one primary error category to invalid instances. Agreement on the binary validity decision is 98.0\%, with Cohen's $\kappa=0.789$. Among the 94 candidates judged invalid by both primary annotators before adjudication, agreement on the primary error category is 90.4\%, with multiclass $\kappa=0.875$.
After adjudication, 96 candidates are confirmed as correctly rejected, while four are judged valid. Table~\ref{tab:mg_rejected_error_analysis} summarizes the adjudicated outcomes. Evidence-grounding failures and ambiguity are the most frequent reasons for justified rejection. The 4.0\% figure refers specifically to valid instances within the sampled rejected pool, rather than to the proportion of all valid candidates rejected by the pipeline.

\begin{table}[t]
\centering
\caption{Human error analysis of 100 rejected Memory-Gym candidates. Each invalid candidate is assigned one primary error category. Counts use adjudicated labels, and percentages are calculated over all 100 sampled rejected candidates.}
\label{tab:mg_rejected_error_analysis}
\small
\setlength{\tabcolsep}{6pt}
\begin{tabular}{@{}lrr@{}}
\toprule
Outcome / primary error type & Count & Percentage (\%) \\
\midrule
Evidence-grounding failure & 33 & 33.0 \\
Ambiguity/non-uniqueness & 22 & 22.0 \\
Query leakage or construction issue & 16 & 16.0 \\
Task-consistency failure & 14 & 14.0 \\
Coverage/completeness failure & 11 & 11.0 \\
\midrule
Correctly rejected (subtotal) & 96 & 96.0 \\
Incorrectly rejected (valid candidates) & 4 & 4.0 \\
\midrule
Total & 100 & 100.0 \\
\bottomrule
\end{tabular}
\end{table}

Together, these audits provide human evidence for the quality of retained Memory-Gym instances and the reliability of automatic filtering on the evaluated samples.

\section{Experiment details}
\label{app:experiment_details}

\subsection{Implementation Details}

\label{app:training_implementation}
All experiments were conducted on computational nodes equipped with 8 NVIDIA H100 80GB GPUs. 
For SFT, we optimize the Researcher on verified trajectories with the standard next-token prediction objective. 
We use Qwen3.5-122B-A10B~\cite{qwen3.5} to generate interaction trajectories and Gemini 3 Flash~\cite{google2025gemini3flash} to filter low-quality or unsupported trajectories. 
A typical SFT run takes approximately 5 hours under this hardware configuration.

Unless otherwise specified, the Researcher performs at most 20 action rounds for each query. 
For each search action, we retrieve the top-5 candidates from BM25 and the top-5 candidates from BGE-M3, and then merge and deduplicate them. 
The browse model and the final answer generation model are both instantiated with the untrained Qwen3.5-4B backbone, and are not updated during SFT or RL.

For RL, we optimize the Researcher policy using Hint-guided GRPO, initialized from the SFT checkpoint and guided by the source-recall reward. 
To improve training efficiency, we use the Verl library~\cite{verl} with vLLM-based rollouts, gradient checkpointing, and FSDP offloading. 
We sample $G=8$ trajectories per query for group-relative advantage estimation and train the policy for 200 optimization steps. 
The policy learning rate is set to $1\text{e-}6$, the KL loss coefficient $\beta$ is fixed at $0.001$, and the clipping ratio $\epsilon$ is set to $0.2$. 
Hints are used only during training rollouts and are removed at test time. 
A typical GRPO run takes approximately 6 hours under the same hardware configuration.

\subsection{LongCodeQA: Cross-Domain Evaluation}
\label{app:longcodeqa_transfer}

We evaluate cross-domain transfer on 233 LongCodeQA queries, comprising
76, 92, and 65 examples in the 64K, 128K, and 256K subsets, respectively.
Memory-Gym contains no code repositories or code-specific tasks, and no
LongCodeQA examples are used for training or adaptation. We compare the
untrained Qwen3.5-4B Researcher, the checkpoint after verified-trajectory
SFT, and the full checkpoint after SFT followed by Hint-guided GRPO;
hints are used only during training. Accuracy is reported for each
context-length subset, while the overall score is computed over all 233
queries rather than as the unweighted mean of the three subset scores.

\subsection{Efficiency Evaluation Details}
\label{app:efficiency_details}

\begin{figure}[t]
    \centering
    \includegraphics[width=0.45\textwidth]{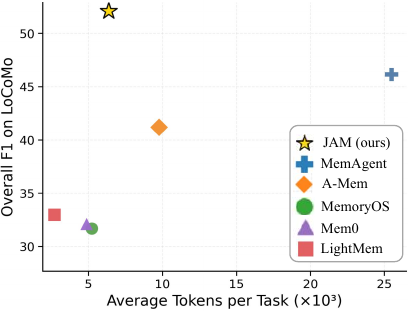} 
    \caption{Efficiency comparison on LoCoMo in terms of average token consumption per task and overall F1 score.} 
    \label{fig:token_efficiency_appendix} 
\end{figure}

\begin{table}[t]
    \centering
    \small
    \setlength{\tabcolsep}{5pt}
    \renewcommand{\arraystretch}{1.08}
    \begin{tabular}{@{}lrrrrrr@{}}
        \toprule
        Setting
        & \shortstack{Offline\\Tokens}
        & \shortstack{Offline\\Time}
        & \shortstack{Online\\Tokens}
        & \shortstack{Online\\Time}
        & \shortstack{Avg.\\Rounds}
        & F1 \\
        & \shortstack{(/workspace)}
        & \shortstack{(s/workspace)}
        & \shortstack{(/query)}
        & \shortstack{(s/query)}
        &  &  \\
        \midrule
        Flat Raw Store
        & 0
        & 0
        & 8174.29
        & 17.11
        & 3.57
        & 49.06 \\
        
        JAM Workspace
        & 36924.51
        & 79.53
        & \textbf{6720.61}
        & \textbf{13.81}
        & \textbf{3.16}
        & \textbf{52.09} \\
        \bottomrule
    \end{tabular}
    \caption{
    Effect of hierarchical workspace organization on LoCoMo.
    The Flat Raw Store is the \textit{w/o Memorizer} variant in
    Table~\ref{tab:ablation}. Both settings preserve the same raw
    histories and use the same trained Researcher, retrievers, answer
    model, and action budget; the Flat Raw Store removes memos,
    hierarchical organization, and README navigation.
    }
    \label{tab:workspace_efficiency}
\end{table}

We evaluate efficiency on LoCoMo under the same system configurations as
the main experiments, separately accounting for one-time offline memory
construction and per-query online serving.

\noindent\textbf{Wall-clock latency.}
We report offline construction time in seconds per workspace and online
serving time in seconds per query, averaged over the LoCoMo evaluation
instances. For JAM, offline time covers hierarchical workspace construction
by the Memorizer, while online time includes the Researcher's iterative
exploration and final query processing.

\noindent\textbf{Token consumption.}
As a hardware-agnostic measure of model-processing cost, we also report
token consumption for offline construction and online serving. For the
per-task comparison in Figure~\ref{fig:token_efficiency_appendix}, the
one-time offline cost is amortized over the queries that reuse the
corresponding workspace and added to the average online token cost per
query. Figure~\ref{fig:token_efficiency_appendix} shows a similar
quality--cost trade-off to the wall-clock results: JAM achieves the
highest LoCoMo F1 while using substantially fewer tokens per task than
MemAgent, whereas the one-shot memory systems incur lower token costs
but also substantially lower answer quality.

\noindent\textbf{Effect of workspace organization.}
We further analyze the \textit{w/o Memorizer} variant from
Table~\ref{tab:ablation}, corresponding to a \textit{Flat Raw Store}
that preserves the same raw histories but removes hierarchical organization
and navigational summaries.

As shown in Table~\ref{tab:workspace_efficiency}, the JAM Workspace incurs
a one-time construction cost of 36,924.51 tokens and 79.53 seconds per
workspace. Once constructed, however, it reduces online token consumption
from 8,174.29 to 6,720.61 tokens per query, latency from 17.11 to
13.81 seconds per query, and average research rounds from 3.57 to 3.16,
while improving LoCoMo F1 from 49.06 to 52.09.

This controlled comparison shows that hierarchical organization does more
than improve answer quality: by providing compact navigational structure
over the same underlying histories, it reduces the amount of online
exploration required by the Researcher.

\subsection{Researcher Runtime and Termination}
\label{app:runtime_behavior}

We characterize the runtime behavior of the fully trained JAM Researcher
across the four main benchmarks under the default maximum budget of
20 action rounds. We report the distribution of exploration length and
the frequency of budget exhaustion.

\noindent\textbf{Round counting and stopping criteria.}
An action round corresponds to one iteration of the Researcher's
thinking--exploration--reflection loop and may contain multiple tool calls;
the subsequent finalization step is not counted as an exploration round.
The Researcher terminates when it produces a positive sufficiency
assessment or reaches the 20-round budget. We report the mean, median,
90th percentile (P90), and proportion of queries stopping within five
rounds. Queries that reach Round 20 without a positive sufficiency
assessment are considered budget-exhausted and are counted as 20 rounds
in the reported statistics.

\begin{table}[t]
    \centering
    \small
    \setlength{\tabcolsep}{6pt}
    \renewcommand{\arraystretch}{1.08}
    \begin{tabular}{@{}lrrrrr@{}}
        \toprule
        Benchmark
        & \shortstack{Stopped within\\5 rounds (\%)}
        & Mean
        & Median
        & P90
        & \shortstack{Budget\\exhausted (\%)} \\
        \midrule
        LoCoMo
        & 95.58 & 3.16 & 3 & 4  & 0.00 \\
        LongMemEval
        & 82.80 & 3.82 & 3 & 6  & 0.40 \\
        NarrativeQA
        & 84.00 & 3.92 & 3 & 8  & 0.00 \\
        HotpotQA
        & 43.83 & 7.11 & 6 & 14 & 4.70 \\
        \bottomrule
    \end{tabular}
    \caption{
        Observed Researcher action rounds under a maximum budget of 20.
        Mean, median, and P90 are computed over all queries, with
        budget-exhausted queries counted as 20 rounds.
    }
    \label{tab:researcher_runtime}
\end{table}

\noindent\textbf{Observed runtime behavior.}
Table~\ref{tab:researcher_runtime} shows that at least 82.80\% of queries
terminate within five rounds on LoCoMo, LongMemEval, and NarrativeQA.
HotpotQA requires longer exploration, with only 43.83\% terminating
within five rounds and a P90 of 14 rounds, compared with 4--8 on the
other benchmarks. Budget exhaustion remains rare, ranging from 0\% to
4.70\% across benchmarks. These results show that the 20-round setting
acts primarily as exploration headroom rather than a fixed inference cost.

\noindent\textbf{Budget-exhaustion handling.}
When the exploration budget is exhausted, JAM applies the same
finalization step used after sufficiency-based termination to construct
a context from the evidence collected so far. The 20-round cap therefore
bounds exploration cost but does not imply that an exhausted query is
incorrect; likewise, a positive sufficiency assessment is an internal
stopping decision rather than a guarantee of complete evidence recovery.

\subsection{Inference-Time Variability}
\label{app:inference_variance}

To assess the sensitivity of JAM to stochasticity during runtime
exploration, we repeat inference on all four main benchmarks using three
independent random seeds. The trained checkpoint, workspace, prompts,
retrieval configuration, action budget, and all other evaluation settings
are held fixed, so only inference-time randomness varies across runs.

\begin{table}[t]
    \centering
    \small
    \setlength{\tabcolsep}{6pt}
    \renewcommand{\arraystretch}{1.05}
    \begin{tabular}{@{}lrrrrr@{}}
        \toprule
        Benchmark & Run 1 & Run 2 & Run 3 & Mean & Std. \\
        \midrule
        LoCoMo F1
        & 52.09 & 52.68 & 51.10 & 51.96 & 0.80 \\
        LongMemEval Acc.
        & 65.20 & 64.60 & 65.00 & 64.93 & 0.31 \\
        NarrativeQA F1
        & 45.72 & 43.37 & 45.21 & 44.77 & 1.24 \\
        HotpotQA F1
        & 60.84 & 58.63 & 58.42 & 59.30 & 1.34 \\
        \bottomrule
    \end{tabular}
    \caption{
    Performance across three independent inference runs using the same
    trained JAM checkpoint. Mean and sample standard deviation are
    reported across inference seeds.
    }
    \label{tab:inference_variance}
\end{table}

Across the four benchmarks, the sample standard deviation ranges from
0.31 to 1.34 points. Variability is lowest on LongMemEval and LoCoMo
and somewhat higher on NarrativeQA and HotpotQA. Overall, the results
remain reasonably stable across inference seeds, with standard deviation
below 1.5 points on all four benchmarks.

\subsection{LLM-as-a-Judge Evaluation}
\label{app:llm_judge}

To assess whether JAM's answer-quality gains persist beyond token-level
F1, we additionally evaluate LoCoMo, NarrativeQA, and HotpotQA using an
LLM judge. For each question, GPT-4o-mini with temperature 0 receives the
reference answer and system prediction and returns \texttt{CORRECT} or
\texttt{WRONG}; accuracy is the proportion of predictions judged
\texttt{CORRECT}. We follow the Memory-R1 evaluation prompt for LoCoMo
and adapt only the task description for NarrativeQA and HotpotQA.
LongMemEval is not re-evaluated because its main metric is already
accuracy. Memory-R1-PPO and Memory-R1-GRPO results are reproduced from
the original paper and are shown separately for reference, rather than
as directly comparable results under our evaluation pipeline.

\begin{table}[t]
    \centering
    \small
    \setlength{\tabcolsep}{9pt}
    \renewcommand{\arraystretch}{1.05}
    \begin{tabular}{@{}lrrr@{}}
        \toprule
        Method & LoCoMo & NarrativeQA & HotpotQA \\
        \midrule
        \multicolumn{4}{@{}l}{\textit{Evaluated with our judge}} \\
        \addlinespace[2pt]
        RAG
        & 58.25 & 48.00 & 53.13 \\
        A-MEM
        & 54.74 & 45.00 & 35.68 \\
        Mem0
        & 46.23 & 43.00 & 38.02 \\
        MemoryOS
        & 47.66 & 40.00 & 30.21 \\
        LightMem
        & 48.57 & 40.00 & 41.15 \\
        MEM1
        & 39.22 & 37.00 & 40.36 \\
        MemAgent
        & 66.62 & 38.00 & 58.59 \\
        \textbf{JAM}
        & \textbf{79.16} & \textbf{68.00} & \textbf{69.79} \\
        \midrule
        \multicolumn{4}{@{}l}{\textit{Reported in prior work}} \\
        \addlinespace[2pt]
        Memory-R1-PPO
        & 57.54 & -- & -- \\
        Memory-R1-GRPO
        & 62.74 & -- & -- \\
        \bottomrule
    \end{tabular}
    \caption{
    LLM-judge accuracy (\%) on three benchmarks.
    The upper block is evaluated using GPT-4o-mini.
    Memory-R1 results are reproduced from the original paper and are
    included only for reference because they are not re-evaluated under
    our judging pipeline. Dashes denote unreported results.
    }
    \label{tab:llm_judge_accuracy}
\end{table}

As shown in Table~\ref{tab:llm_judge_accuracy}, JAM achieves the highest
accuracy among methods evaluated under the same judging protocol on all
three benchmarks. Compared with the strongest baseline in this group,
JAM improves accuracy by 12.54, 20.00, and 11.20 percentage points on
LoCoMo, NarrativeQA, and HotpotQA, respectively. These results show that
JAM's gains persist under a complementary answer-level evaluation,
providing additional evidence that the improvements are not limited to
lexical-overlap-based scoring.

\subsection{Memorizer Backbone Sensitivity}
\label{app:memorizer_backbone}

We examine the sensitivity of JAM to the Memorizer backbone on LoCoMo
by replacing the default Qwen3.5-4B Memorizer with Qwen3.5-122B-A10B
and GPT-5.5. Each backbone constructs its own workspace from the same
raw histories, while the trained Qwen3.5-4B Researcher, retrievers,
browse and final-answer models, action budget, and other online inference
settings are held fixed. The Researcher is not retrained for the
alternative workspaces.

\begin{table}[t]
    \centering
    \small
    \setlength{\tabcolsep}{6pt}
    \renewcommand{\arraystretch}{1.08}
    \begin{tabular}{@{}lrrrr@{}}
        \toprule
        Memorizer backbone
        & \shortstack{Build time\\(s/workspace)}
        & \shortstack{Online time\\(s/query)}
        & \shortstack{Avg.\\rounds}
        & F1 \\
        \midrule
        Qwen3.5-4B (default)
        & 79.53 & 13.81 & 3.16 & 52.09 \\
        Qwen3.5-122B-A10B
        & 132.43 & 14.36 & 3.25 & 51.30 \\
        GPT-5.5
        & 117.65 & 13.19 & 3.03 & 52.62 \\
        \bottomrule
    \end{tabular}
    \caption{
    Effect of the Memorizer backbone on LoCoMo.
    Each backbone constructs its own workspace from the same raw
    histories, while the trained Researcher and online inference
    configuration are held fixed.
    }
    \label{tab:memorizer_backbone}
\end{table}

As shown in Table~\ref{tab:memorizer_backbone}, changing the Memorizer
backbone does not yield a consistent improvement in downstream performance.
Qwen3.5-122B-A10B decreases F1 from 52.09 to 51.30, whereas GPT-5.5
increases it modestly to 52.62. Both alternatives incur higher workspace
construction cost, while online latency and average exploration rounds
change only slightly. The default Qwen3.5-4B Memorizer therefore provides
a favorable balance between workspace-construction cost and downstream
performance in the evaluated setting.

We keep the Memorizer fixed during Researcher training because its
query-agnostic workspace is reused across requests, providing a stable
environment for learning query-conditioned exploration. We do not study
Memorizer fine-tuning or joint Memorizer--Researcher optimization; this
experiment isolates sensitivity to the Memorizer model choice.

\section{Case Study}

\begin{figure}[htbp]
    \centering
    \includegraphics[width=\textwidth]{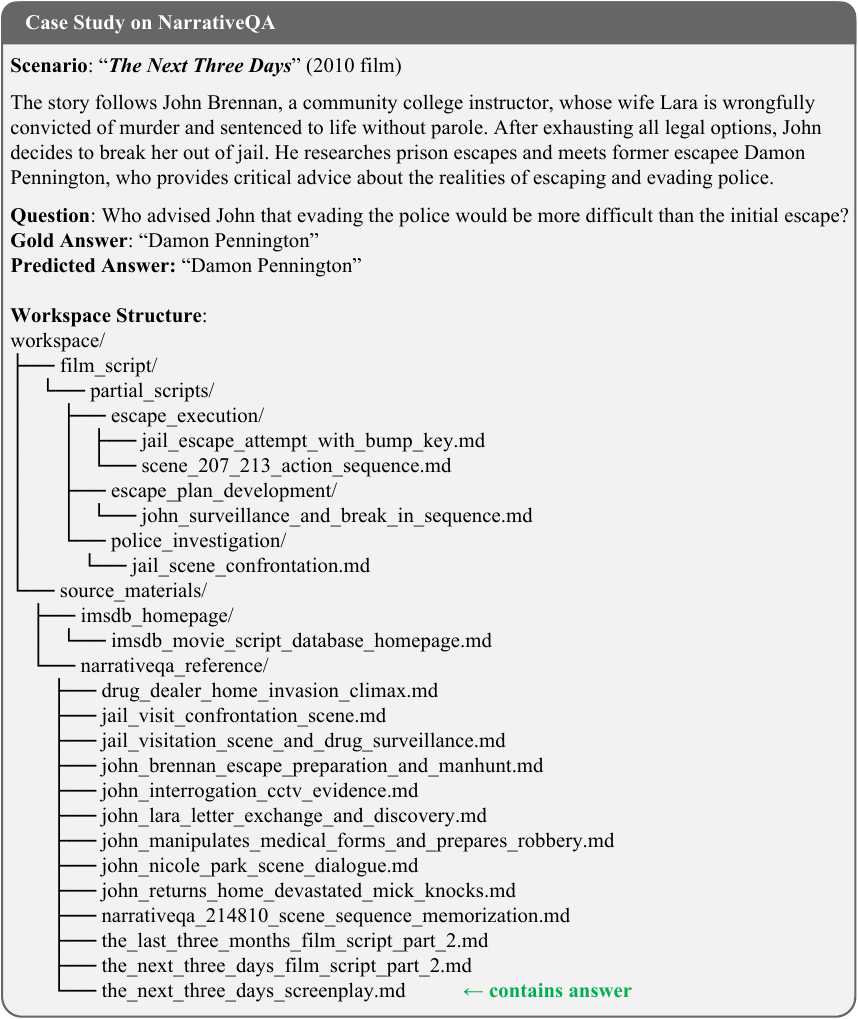} 
    \caption{Case-study setup and Memorizer-constructed workspace for a NarrativeQA example.} 
    \label{fig:case_study_1} 
\end{figure}

\begin{figure}[htbp]
    \centering
    \includegraphics[width=\textwidth]{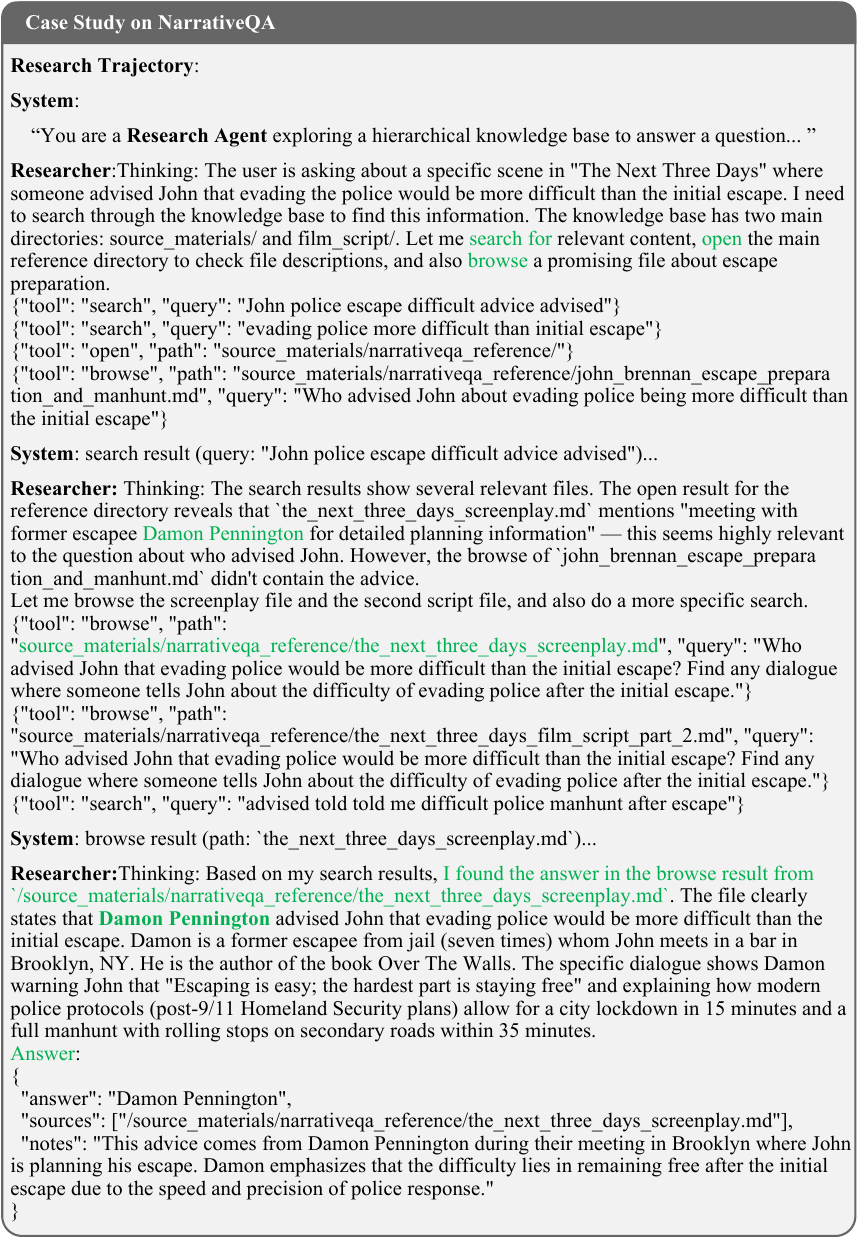} 
    \caption{Researcher trajectory for locating the supporting evidence in the NarrativeQA case.} 
    \label{fig:case_study_2} 
\end{figure}

To clearly illustrate the execution process of the Researcher, we present a representative NarrativeQA case in Figure~\ref{fig:case_study_1} and Figure~\ref{fig:case_study_2}. 
As shown in Figure~\ref{fig:case_study_1}, the Memorizer converts the original narrative materials into a structured workspace with two major branches, \texttt{film\_script/} and \texttt{source\_materials/}. 
This structure preserves raw evidence while exposing navigable paths, allowing the Researcher to reason over the workspace at both coarse and fine granularities.

Figure~\ref{fig:case_study_2} shows the corresponding research trajectory. 
After receiving the query, the Researcher first inspects the overall workspace and issues multiple complementary tool calls, including \texttt{search}, \texttt{open}, and \texttt{browse}, to explore potentially relevant regions. 
After observing the initial results, it identifies a more promising evidence path and performs focused browsing over the screenplay file. 
This coarse-to-fine trajectory allows the Researcher to progressively narrow the search space, locate the supporting evidence, and return the correct answer with source provenance. 
The case illustrates that JAM enables the Researcher to start from global workspace understanding, combine multiple memory-access tools for efficient exploration, and incrementally construct query-relevant evidence beyond one-shot retrieval.

\section{Prompt}
\label{app:prompt}
Table~\ref{tab:researcher_prompt} shows the prompt template used by the Researcher for memory exploration. 
The prompt specifies the Researcher's role, the available tools, the think--tool-use loop, and the required source-grounded answer format.



{
\setlength{\LTpre}{0pt}
\setlength{\LTpost}{0pt}
\setlength{\abovecaptionskip}{2pt}
\setlength{\belowcaptionskip}{3pt}

\begin{longtable}{@{}p{0.98\linewidth}@{}}
\caption{\normalsize Researcher prompt for deep-research-style exploration.}
\label{tab:researcher_prompt}\\[-2pt]
\toprule
\textbf{Researcher Prompt Template} \\
\midrule
\endfirsthead

\bottomrule
\endfoot

\begin{minipage}[t]{\linewidth}
\fontsize{7.7pt}{8.7pt}\selectfont
\setlength{\parindent}{0pt}
\setlength{\parskip}{0pt}

\textbf{Your Role.}\\
You are a \textbf{Research Agent} exploring a hierarchical knowledge base to answer a question.\\[1pt]

\textbf{Knowledge Base Structure.}\\
A \textbf{Knowledge Base Overview} is provided at the end of this system prompt. It includes a summary of the knowledge base content and the full directory structure. Each folder has a README that summarizes the files and subfolders inside it. You can use these README files to quickly understand the workspace structure and decide which files to inspect.\\[1pt]

\textbf{Available Tools.}\\
\textbf{1. \texttt{search}.} Search for files matching a query in the knowledge base. It returns relevant file names, paths, and summaries. Use this tool to find potentially relevant files. Parameter: \texttt{query}, a keyword or phrase to search for in file content.\\
\textbf{2. \texttt{browse}.} Open a file and extract information relevant to a query. An AI assistant reads the file and returns a summary of query-relevant content. Parameters: \texttt{path}, the file path to browse; \texttt{query}, a specific query asking for the exact information needed from the file.\\
\textbf{3. \texttt{open}.} Open a folder to view its README summary and directory listing. Use this tool to understand the contents and structure of a directory. Parameter: \texttt{path}, the folder path to open.\\[1pt]

\textbf{Process.}\\
Follow the think $\rightarrow$ tool-use loop.
(1) Think about what is known, what is missing, and which tool should be used.
(2) Use one or more tools to gather information, with at most five tool calls per round.
(3) Receive observations from the tools.
(4) Think again based on the observations and decide the next step.
(5) Repeat until enough information has been collected.
(6) Output the final answer in \texttt{\textless answer\textgreater} tags.\\[1pt]

\textbf{Output Format During Exploration.}\\
Each exploration round should contain one \texttt{\textless think\textgreater} block followed by one or more \texttt{\textless tool\_use\textgreater} blocks. Each tool call must be wrapped in its own \texttt{\textless tool\_use\textgreater} block and must contain valid JSON.\\[-1pt]
\texttt{\textless think\textgreater}\\
\textit{[Reason about what you know, what information is missing, which tool to use, and why.]}\\
\texttt{\textless /think\textgreater}\\
\texttt{\textless tool\_use\textgreater}\\
\texttt{\{"tool": "search", "query": "your search query"\}}\\
\texttt{\textless /tool\_use\textgreater}\\
\texttt{\textless tool\_use\textgreater}\\
\texttt{\{"tool": "browse", "path": "your browse path", "query": "your browse query"\}}\\
\texttt{\textless /tool\_use\textgreater}\\[1pt]

\textbf{Final Answer Format.}\\
When enough information has been collected, output a final answer in the following format and stop the exploration loop.\\[-1pt]
\texttt{\textless think\textgreater}\\
\textit{[Summarize the collected evidence and reasoning.]}\\
\texttt{\textless /think\textgreater}\\
\texttt{\textless answer\textgreater}\\
\texttt{\{"answer": "Your comprehensive answer to the question", "sources": ["/path/to/source1.md", "/path/to/source2.md"], "notes": "Additional notes or caveats"\}}\\
\texttt{\textless /answer\textgreater}\\[1pt]

\textbf{Guidelines.}\\
First check the Knowledge Base Overview, since it already provides the directory tree.
Use multiple tools in one round when this can speed up exploration, but use at most five tool calls per round.
Use multiple rounds of thinking and tool use until the evidence is sufficient.
When outputting \texttt{\textless answer\textgreater}, make sure the answer is grounded in collected evidence and includes source paths.
Each \texttt{\textless tool\_use\textgreater} block must contain valid JSON with a \texttt{"tool"} field.\\[1pt]

\textbf{Begin.}\\
Review the Knowledge Base Overview, identify the most relevant areas for the question, and start exploring.\\[1pt]

\textbf{Knowledge Base Overview:} \textit{\{knowledge\_base\_overview\}}

\end{minipage}
\\

\end{longtable}
}

\end{document}